# Unified Pix Token And Word Token Generative Language Model

**Haun Leung** haunleung@outlook.com
**ZiNan Wang** wangzinan@buaa.edu.cn

## Abstract

Since the emergence of Vision Transformer (ViT), it has been widely used in generative language model and generative visual model. Especially in the current state-of-art open source multimodal models, ViT obtained by CLIP or SigLIP method serves as the vision encoder backbone to help them acquire visual understanding capabilities. But this method leads to limitations in visual understanding for details, such as difficulty in recognizing small text or numbers in images. To address these issues, we propose a new model to unify pix token and word token into the generative language model. The new model also features with each pix of image having its own token embedding, color folding, global conditional attention approximation and image unsupervised pretraining. We conducted image unsupervised pretraining experiments using our new model to explore its potential. The experimental results show that it has good performance even in small model and with limited training data. We believe our model also conforms to the scaling law, as long as model parameters and training data increased, its performance will continue to improve.

## 1 Introduction

Since the emergence of Vision Transformer (ViT), it has been widely used in generative language model and generative visual model. Especially in the current state-of-art open source multimodal models, just like Llava, ViT obtained by CLIP or SigLIP method serves as the vision encoder backbone to help them acquire visual understanding capabilities. But this method leads to some issues. Either there are limitations in visual understanding for details, such as difficulty in recognizing small text or numbers in image. Either there are visual understanding drifts in multi-turns of conversations, the first turn of conversation is correct, but errors are prone to occur later on. Although these limitations can be alleviated by adding more training data related to the issues or optimization CLIP method, there will still be other challenges in the future.

We believe that these limitations are likely due to the CLIP method, including the following reasons.
**1) The encoded features cannot represent all the details of the image**
ViT encodes images into a small features that is similar to a text features. But what if this text does not fully describe other small details of the image? Or can the text describe all the details of the image? In this case, the encoded features of the image cannot represent all the details of the image.

**2) The slightest nudge causes the widest chain reaction**
If a small change is made for the image, for example the license plate number of car in the image changes from 1 to 6, the all floating-point numbers in the encoded features will change. The slightest nudge causes the widest chain reaction, which is not good.

**3) ViT use a fake token embedding**
ViT forms a fake token embedding by patching the pixels in the image. But this fake token embedding does not have truly adjustable parameters. So the expressive ability of ViT is questionable. We may need to propose a new architecture to address its shortcomings.

Table 1: Model Configuration Information

| dim | layers | heads | kv heads | image dim | image layers | fold factor | image size | window size |
|---|---|---|---|---|---|---|---|---|
| 768 | 12 | 12 | 6 | 768 | 5 | 16 | 224 | 16 |

**4) Limitations of supervised training**
CLIP uses supervised training, but the supervised training data is always limited. It is easy to cause deficiencies outside of distribution.

We creatively proposed a new architecture to unify pix token and word token into the generative language model, no longer relying on ViT encoder. We believe that each pix of image should have its own unique token and true token embedding that we call Pix Token Embedding. We also proposed the concept of Color Folding, which reduces the huge amount of computation and makes Pix Token Embedding feasible. Furthermore, we have proposed Global Conditional Attention Approximation that enables the entire lengthy Pix Token Embedding sequence of image to participate in self-attention operations effectively and feasibly. Now the new architecture make generative language model can predict pix token like predicting word token. We also proposed image unsupervised pretraining, which enable model to learn the patterns of the pixes from huge image datas. And our method is lossless in image data preprocessing, but models like ViT suffer losses in image data preprocessing.

We conducted experiments on image unsupervised pretraining, using our unified pix Token and word token model. We found that it has the potential to learn the patterns of the pixes. The experimental results show that it has good performance even in small model and with limited training data. We believe our architecture also conforms to the scaling law, as long as model parameters and training data increased, its performance will continue to improve.

However, due to our limited computing resources, we are unable to conduct experiments with large-scale model and data. We cannot predict the upper limit of its potential.

God that creates colorful, please please blesses us with some computing resources. We hope our research results [1] can benefit the whole humanity you created.

## 2 Related work

### 2.1 ViT (Vision Transformer) and application

**ViT (Vision Transformer) and application** [Dosovitskiy et al., 2020] is a computer vision model based on Transformer architecture [Vaswani et al., 2017]. It transforms image processing tasks into sequence processing problems, making breakthrough progress in multiple visual tasks. The core idea of ViT is to view an image as a sequence of image patches. Specifically, it devides the input image into fixed sized non overlapping small patches, and then linearly projects each small patch into a token embedding. These token embedding sequences, along with an additional learnable classification token embedding, form the input of the Transformer encoder.

**CLIP (Contrastive Language-Image Pre-training)** [Radford et al., 2021] is a multimodal visual language pre training model proposed by OpenAI in 2021. Its core goal is to map images and text to a unified semantic space through contrastive learning, achieving cross modal semantic alignment.

**SigLIP (Sigmoid Loss for Language-Image Pre-training)** [Zhai et al., 2023] is a multimodal pre training model proposed by Google DeepMind in 2023. It is an improved version of CLIP, with the core innovation of using Sigmoid loss function instead of traditional Softmax contrastive loss, significantly improving training efficiency and cross modal alignment performance.

CLIP and SigLIP are multimodal contrastive learning models based on the Vision Transformer (ViT) architecture, aimed at mapping images and text to the same semantic space for cross modal understanding.

In LLM, ViT obtained through CLIP or SigLIP methods is used as the vision encoder backbone to assist large language models in acquiring visual understanding capabilities. For example, LLAVA [Liu et al., 2023]. Currently, excellent open-source multimodal LLMs are further optimized based on the SigLIP method to achieve better image understanding capabilities, such as Kimi-2.5 [Team et al., 2026] and QWen-2.5-VL [Bai et al., 2025].

[1] https://github.com/HaunLeung/upw.git

### 2.2 Unsupervised pretraining

Research work in this area includes GPT-3 [Brown et al., 2020] and so on.

**GPT-3** builds language understanding ability by learning to predict the next word through unsupervised pretraining on massive Internet texts (about 300B tokens). After unsupervised pretraining, the GPT-3 model can adapt to new tasks with a small number of examples or natural language instructions without fine-tuning through few shot learning and zero shot learning capabilities. The success of GPT-3 has also driven the development of subsequent models such as GPT-3.5 and ChatGPT, becoming an important cornerstone of the generative AI revolution. Unsupervised pretraining and supervised posttraining have become the standard pipelines for training generative models.

GPT-3 proves the importance of training with massive text data. CLIP or SigLIP method is only the ViT encoder obtained by training on limited supervised image data. But our model is capable of training on massive unsupervised image data. This laid the foundation for success in visual understanding.

### 2.3 Pix Token and Image Tokenizer

Research work in this area includes iGPT [Chen et al., 2021], image tokenizer like TokenFlow [Qu et al., 2025], MUSE-VL [Xie et al., 2025], BLIP3-o [Chen et al., 2025], EMU2 [Sun et al., 2024], QLIP [Zhao et al., 2025], UniTok [Ma et al., 2025], UniFlow [Yue et al., 2026] and so on.

**iGPT** applies Transformers to image pixels after reducing the image resolution and color space. The model is trained as a generative model in an unsupervised manner and can then be fine-tuned or linearly probed on the obtained representations to achieve classification performance, reaching a maximum accuracy of 72% on ImageNet. In iGPT, the method for handling image resolution and color space involves first using the input resolution (IR) approach to reduce the resolution. Then, by applying k-means with k=512 to cluster the (R, G, B) pixel values, the context size is further reduced by a factor of 3, resulting in the model resolution (MR). [Ranzato et al., 2016] applied a similar method to spatial patches. Ultimately, the VQ-VAE [Delaitre et al., 2011] method was employed to downsample the images after MR. The clustering of (R, G, B) values by iGPT can also be interpreted as training a VQ-VAE using an identity encoder and decoder. iGPT ultimately selected a latent vocabulary size of 4096, which is the minimum size without observed reconstruction artifacts.

**TokenFlow, MUSE-VL** employ pixel and semantic encoders to address generation and understanding tasks, respectively.

**BLIP3-o, EMU2** utilize frozen, well-pretrained vision foundation models as visual encoders and incorporate a latent diffusion decoder for pixel reconstruction.

**QLIP, UniTok** initialize visual encoders using pretrained foundation models and finetune them with a pixel decoder by directly mapping semantic tokens to pixel targets for unification.

**UniFlow** similar to QLIP, UniTok but flexibly adjust any visual encoder by using a concise reconstruction decoder.

**Some differences:** iGPT and the above-mentioned image tokenizer, etc are all rely on Vector Quantization (VQ) method. VQ compresses the entire patch into an index. Similar patchs have the same index, making it difficult to distinguish pix level details.

Our model doesn't need VQ at all. Each pix has a pix token and a pix token embedding and enough to express pix level details. And we use color folding to compress the space of collor and make the total number of pix tokens reasonable. The color folding method has a simple logic and conforms to the laws of the human eye and nature, without the need for complex and meaningless reconstruction.

### 2.4 Global attention with Transformer

The Transformer architecture employing full attention mechanisms on full-sized images requires substantial computational resources, which exceed current capabilities and are thus unfeasible. Current research efforts are exploring ways to reduce computational load or avoid direct full-attention mechanisms. Research work in this area includes Swin Transformer [Liu et al., 2021], QWen-2.5-VL [Bai et al., 2025] and so on.

**Swin Transformer** uses shifted window based self-Attention to avoid full attention Swin Transformer, like ViT, divides images into token embeddings based on patches. Then use multi head self-attention based on shifted window to these token embeddings. Shifted windows include W-MSA and SW-MSA. W-MSA is a regular window, while SW-MSA is an irregular window. Swin Transformer will also merge token embeddings layer by layer.

**Qwen2.5-VL** employs a redesigned Vision Transformer (ViT) architecture as vision encoder to enhance performance and efficiency. Structurally, they incorporate 2D-RoPE and window-based attention mechanisms to support native input resolutions while accelerating the computation of the entire visual encoder. During both training and inference, the height and width of the input images are resized to multiples of 28 before being fed into the ViT. In their architecture, only four layers employ full self-attention, while the remaining layers utilize windowed attention with a maximum window size of 112×112 (corresponding to 8×8 patches). Regions smaller than 112x112 are processed without padding, preserving their original resolution. This design allows the model to operate natively at the input resolution, avoiding unnecessary scaling or distortion. But in the Qwen2.5-VL paper, the architecture figure show that they use M window-based attention blocks in front and the only last block using full attention.

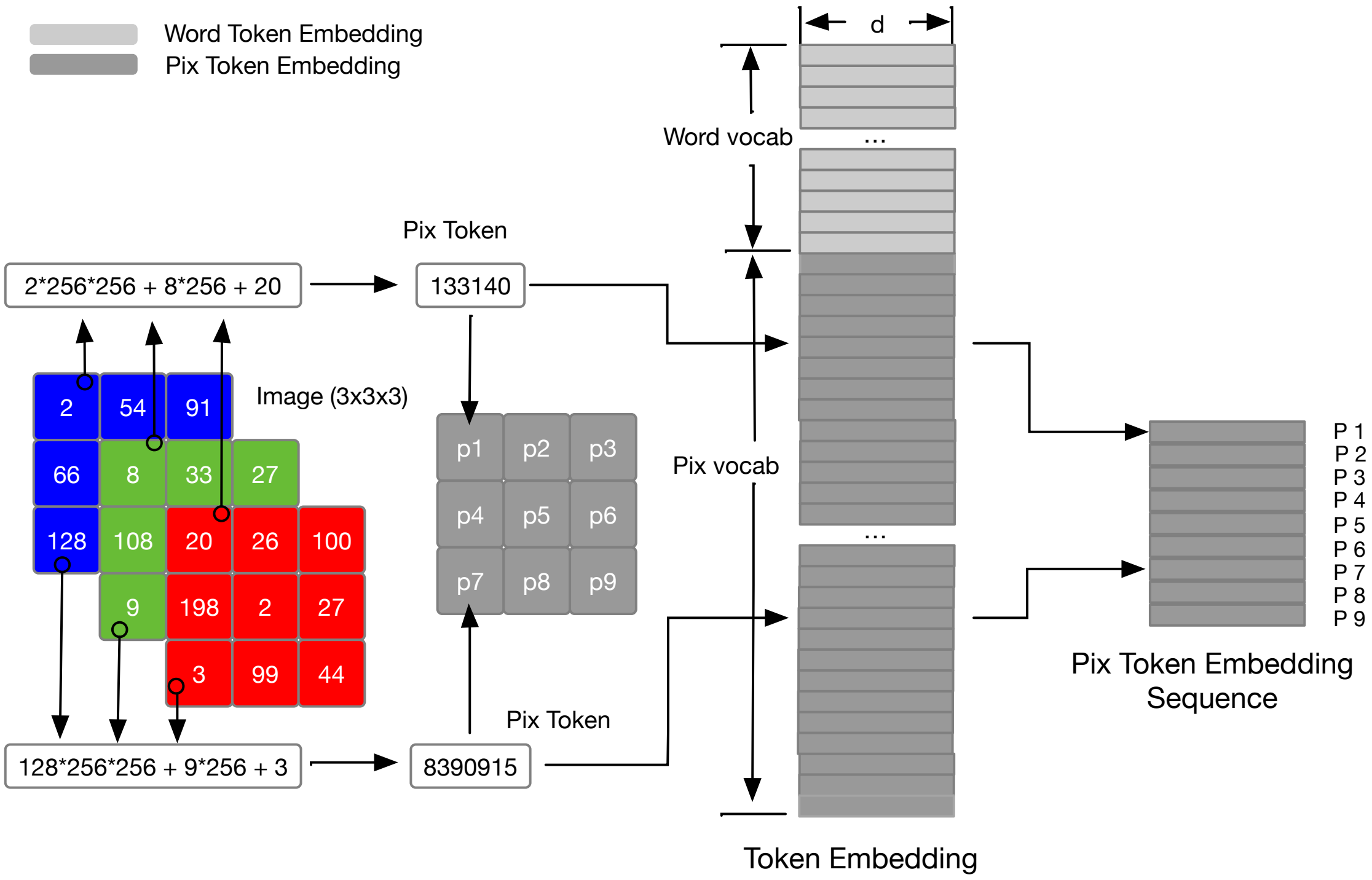

Figure 1: Pix Token and Pix Token Embedding.

**Some differences:** Swin Transformer first uses regular window attention, and then uses inregular window attention. QWen-2.5-VL first used regular window attention, and finally used global attention. QWen-2.5-VL and Swin Transformer both use token embedding merging. Our proposed method is different from Swin Transformer and QWen-2.5-VL in that it does not require complex inregular window attention, full attention, or token embedding merging. It only requires simple local window mask multi-head self-attention operation.

Swin Transformer and QWen-2.5-VL, like ViT, divide images into token embeddings based on patches, which is a type of fake token embeddings without adjustable parameters during training. But we are a true per pix token embedding.

## 3 Method

The unified pix token and word token model we propose is based on Pix Token Embedding, which we will discuss in section 3.1. In addition, to reduce computational complexity, we propose Color Folding, which will be discussed in section 3.2. We will discuss our unified model in section 3.3. At last we discuss its advantages in section 3.4. Section 3.5 is the supplement details during implementation of the model.

### 3.1 Pix Token Embedding

In the current popular LLMs, if we ignore the fancy tokenizing method of the tokenizer, we can approximately assume that each word token has its own independent token embedding. It is this method that gives each word token sufficient expressive power, and after training with massive text data, its parameters can be adjusted to appropriate positions, thereby improving the ability of LLMs.

ViT divides the pixels in the image into patches, and then forms a token embedding by channel for each pixel in each patch. We believe that this approach at least has two drawbacks:
First it does not have adjustable parameters, it is a fake token embedding.
Second not every pix has its own independent token embedding.

Before proposing our method, we make the following assumptions: The shape of image with 3 color channels (RGB channels) is (3, H, W), where 3 is the number of channels, H is the image height, and W is the image width. And the length of the color channel is L which defaults to 256.

We believe that every pix should also have its own true token embedding. To achieve this goal, our design is as follows: Each pix of image corresponds to a unique one token which has one's own token embedding. As shown in Figure 1.

The value of pix token is related to the color value of the pix. The formula for calculating the value of token is:

$$Token_{value} = r + g * L + b * L * L$$

L is the length of the color channel. r is R channel value and g is the G channel value and b is the B channel value.

If L is 256, the token value range is $[0, 256 * 256 * 256]$. So the total number of tokens dictionary is $256 * 256 * 256$.

Because every token have it's own token embedding, so the total number of token embeddings is $256 * 256 * 256$. The dimension of token embedding is set according to the actual situation, and we define it as d. In space, these token embeddings are continuous and the tensor shape is $(256 * 256 * 256$, d).

Attention: Actually also there is an additional pad token for pix due to differences in height and width of image.

### 3.1.1 Pix Token Embeddings Sequence

Next, we will investigate how to obtain a token embeddings sequence from all the pixels in the image. Our combination method is sequential: In order from left to right, and then from top to bottom, extract a pix token from the image and find its corresponding pix token embedding. Then combine these pix token embeddings in chronological order to form a token embeddings sequence.

Taking an image with L=256 and height H and width W both being 3 as an example, as shown in Figure 1:
The first pix $p_1$ is $(20, 8, 2)$, its token value is equal to $2 * 256 * 256 + 8 * 256 + 20 = 133140$.
The 7th pix $p_7$ is $(3, 9, 128)$, its token value is equal to $128 * 256 * 256 + 9 * 256 + 3 = 8390915$.
Similarly, we obtain a pix token sequence $[p_1, p_2, p_3, p_4, p_5, p_6, p_7, p_8, p_9]$.
Then based on this pix token sequence, we get a pix token embedding sequence from the token embedding dictionary.

### 3.1.2 Lossless of Data Preprocessing

As is well known, image data needs to be preprocessed before model training. The traditional method is to map RGB color values from $[0 - 255]$ to $[0 - 1.0]$, for example $1/255 = 0.00392156862745098$. Due to the limited number of floating-point digits in tensors, it is necessary to perform bit truncation on color values, which can result in losses. Although we think the loss value is relatively small, considering the lower learning rate and different optimizer methods and considering training with 16 bits or lower, there are imperfections.

The Pix Token Embedding method we propose will not have this problem, it is lossless data preprocessing.

## 3.2 Color Folding

For an image with 3 color channels (RGB channels), if the length of each channel L is 256, then each pix has $256 * 256 * 256$ colors available, there are over 16.7 million species in total. This quantity is enormous, adding complex calculations. But in reality, we can use less colors to express it. For example, suppose the length of each channel L is 128, then each pix has $128 * 128 * 128$ colors available. In this situation, our human eyes cannot distinguish it, compared to channel length 256.

Use another example to illustrate the feasibility of reducing the variety of colors. Assuming there is a color c0 is $(32, 128, 192)$, form is (R, G, B). We made some changes to the RGB channels value and obtained $c_1$ , $c_2$, $c_3$, $c_4$ four colors. We found that the human eye can hardly distinguish the color details after these changes, as shown in Figure 2.

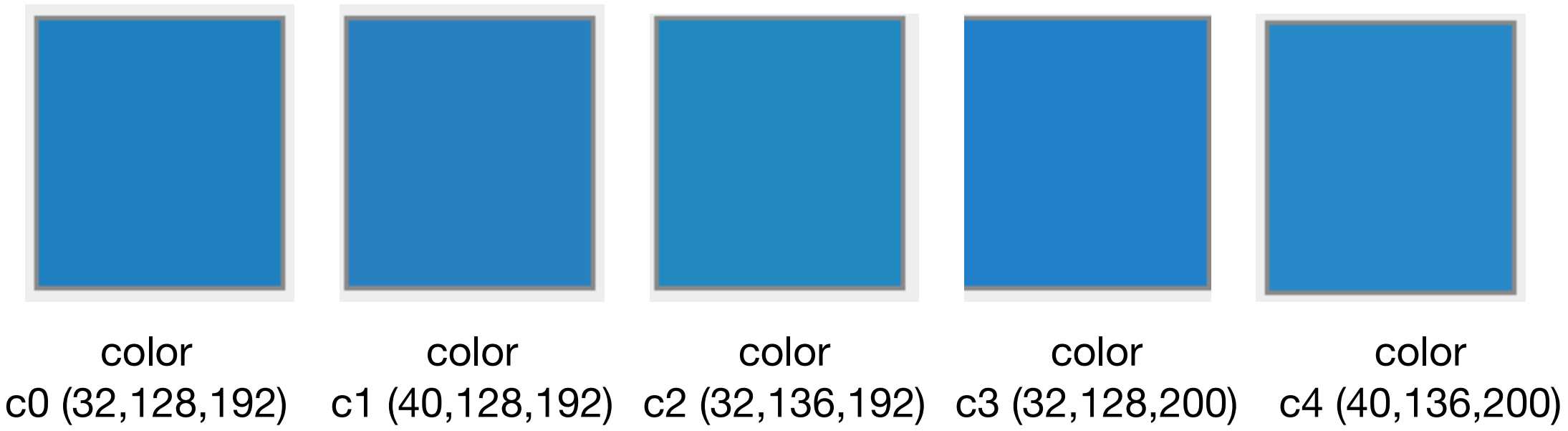

Figure 2: Small changes to the RGB channels value of color.

$c_0$ $(32, 128, 192)$, original color.
$c_1$ $(40, 128, 192)$, R channel value channges 8.
$c_2$ $(32, 136, 192)$, G channel value channges 8.
$c_3$ $(32, 128, 200)$, B channel value channges 8.
$c_4$ $(40, 136, 200)$, RGB channel value channges 8.

Since we humans cannot distinguish it, machines can also do without the ability to distinguish details. Therefore, we can express the physical world with fewer colors, instead of using $256 * 256 * 256$ colors.

#### 3.2.1 Folding Factor

We propose the concept of Color Folding, and its idea is as follows: We propose a Folding Factor whose range of values is $[1, 2, 4, 8, 16, 32, 64, 128, 256]$. Assuming the length of each color channel L is 256, the folded color length is equal to $(256/foldingfactor)$. Folding factor reduces the length of color channel, which is equivalent to fewer color species, which can reduce unnecessary computation.

For example, before folding, each pix has $256 * 256 * 256 = 16777216$ colors. If folding factor is 8, then each pix has $32 * 32 * 32 = 32768$ colors. The reduction of colors is enormous!

Due to the consistency between the total number of pix tokens and the total number of colors, the reduction in computational complexity is also significant.

**Color Value after Color Folding:**
Assuming that the color of a pix is $(R, G, B)$ and folding factor is f. After color folding the color of the pix is $(R//f, G//f, B//f)$. It means the new color value of another color space.

#### 3.2.2 Visualization of Color Folding

Assuming that the color of a pix is $(R, G, B)$ and folding factor is f. After color folding the color of the pix is $(R//f, G//f, B//f)$. For visualization, the color value after unfolding is $((R//f) * f, (G//f) * f, (B//f) * f)$.

Figure 3 is a visual comparison of different folding factor. We can see that for folding factor 2 and 4, our naked eye cannot distinguish them from the original image. For folding factor 8, the difference is very small, almost indistinguishable from the original image. For folding factor 16, there are some minor distortions. For folding factor 32, there is a very obvious distortion.

Color folding can effectively reduce computational complexity while expressing images without distortion. In today's world where computing resources are so precious, this is a feasible approach.

#### 3.2.3 Selection of Folding Factor

The relationship between the total number of pix tokens and folding factor as shown in Table 2. The larger the folding factor, the smaller the total number of pix tokens, and the less computing resources are needed.

The relationship between computing resources and folding factor: In situations where computing resources are limited, you can try using 16. In situations where computing resources are sufficient, you can try using 8. In the case of abundant computing resources, you can try using 4. If it is greater than or equal to 32, it is not recommended because the distortion is very severe. If it is less than 4, it is completely unnecessary, increases the computational load, and the feasibility is not high.

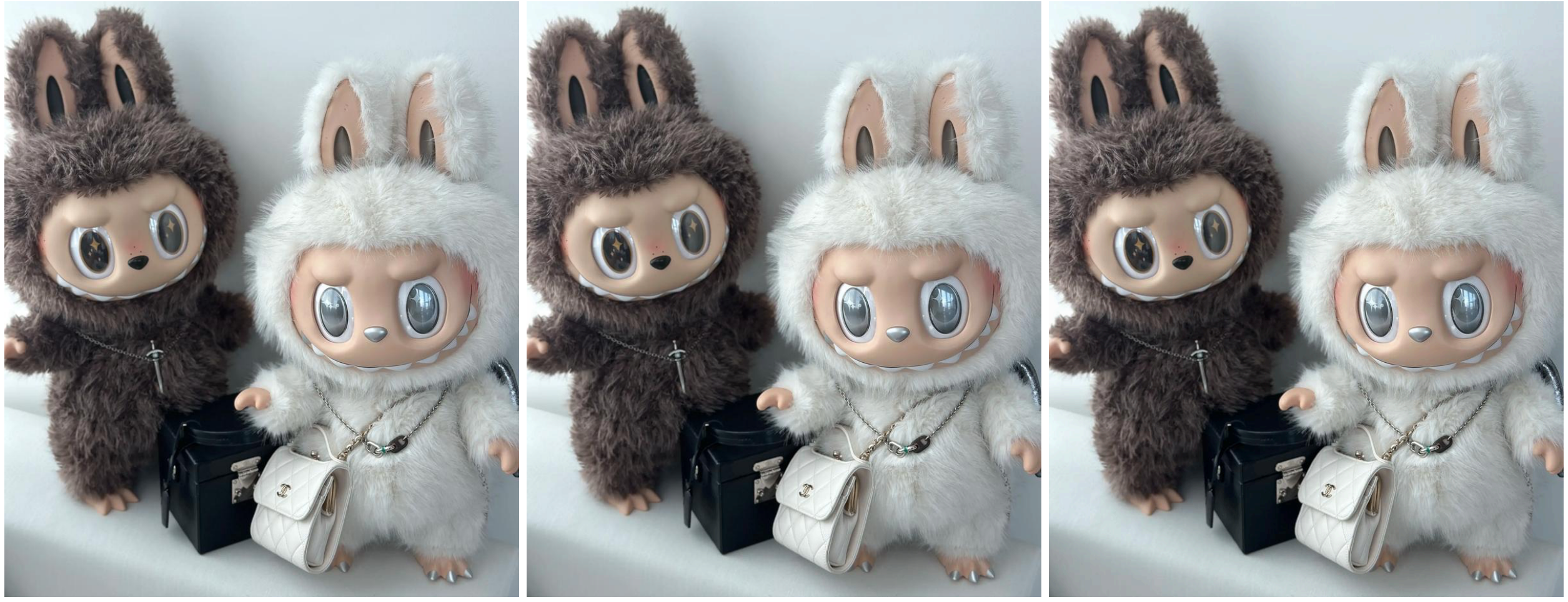

(a) The original (b) Folding factor = 2 (c) Folding factor = 4

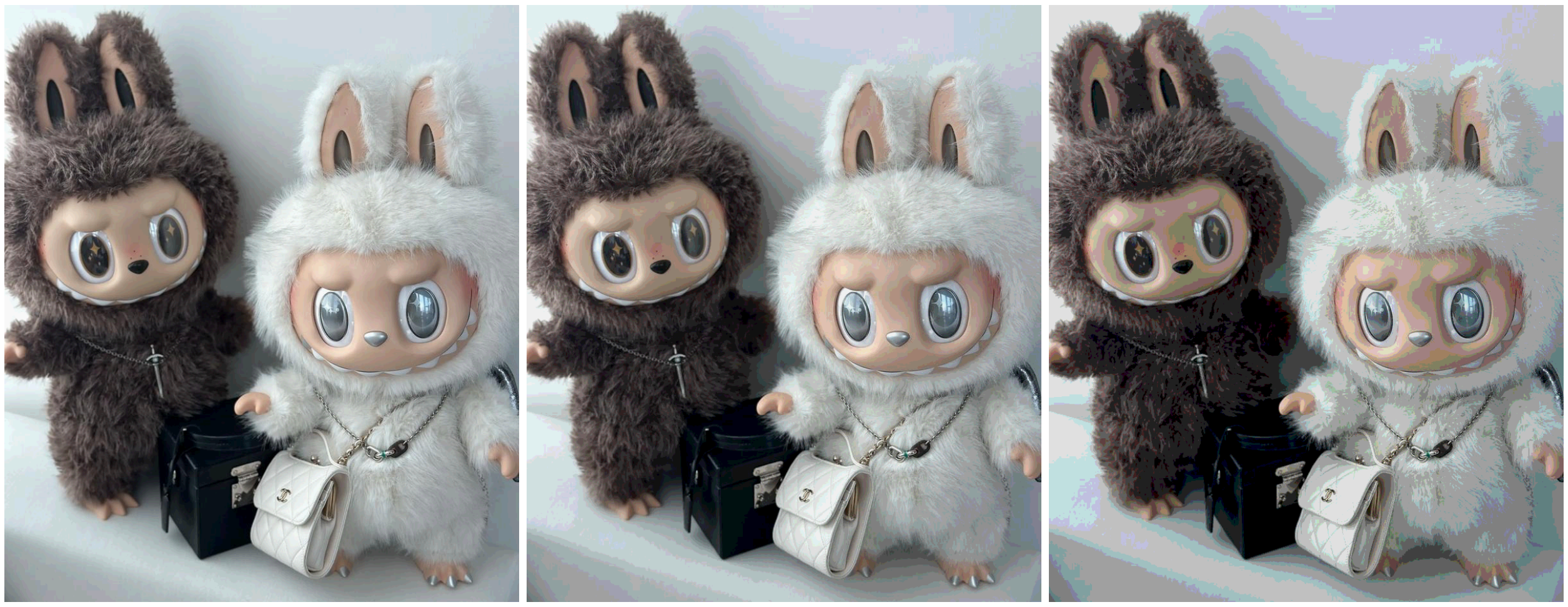

(d) Folding factor = 8 (e) Folding factor = 16 (f) Folding factor = 32

Figure 3: Visualization of Color Folding.

Table 2: Folding Factor and Pix Token Total

| Folding Factor | Pix Token Total |
|---|---|
| 2 | 2097152 |
| 4 | 262144 |
| 8 | 32768 |
| 16 | 4096 |
| 32 | 512 |

### 3.3 Unified Pix Token And Word Token Model

We propose unified pix token and word token generative language model: It unifys the pix token into the generative language model where only word token is used. The core idea of our model are: one is unified pix token and word token, and the other is using local conditional self-attention to approximate global conditional self-attention. It overcomes the drawbacks of large computational complexity and slow speed while approaching the performance of global conditional self-attention. An overview of our model is depicted in Figure 4.

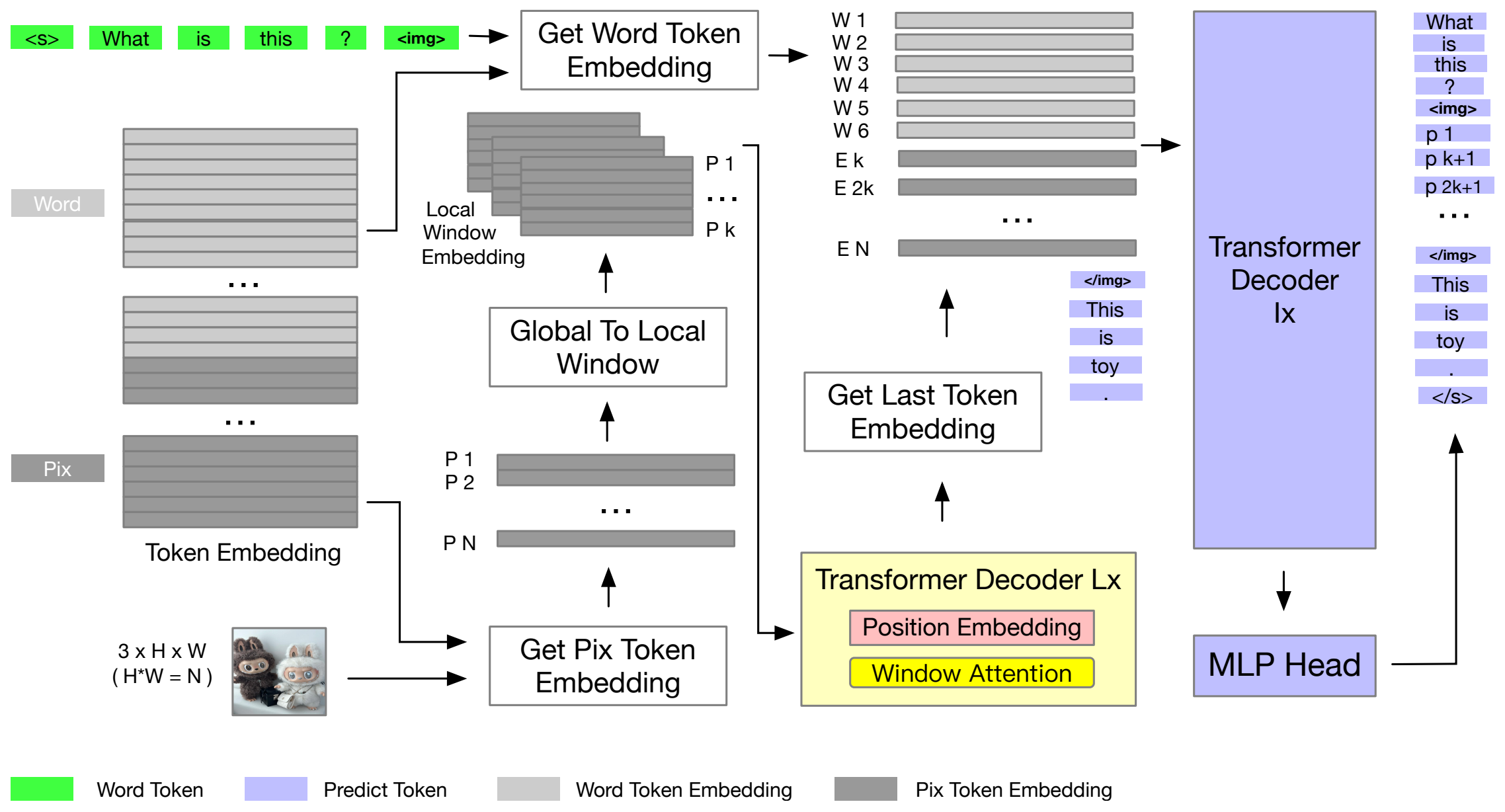

Figure 4: Unified Pix Token And Word Token Model overview.

The model process is as follows:
1) Get Word Token Embedding
For the input model text sequence, we obtain word token embedding sequence based on word token.
2) Get Pix Token Embedding
For the input image (3, H, W), we first obtain a pix token embedding sequence from the token embedding dictionary using the method described in section 3.1.1.
3) Global To Local Window
Convert the global pix token embedding sequence into local window pix token embedding sequence.
4) Local Window Mask Multi-head Self-Attention
Then send the local window pix token embedding sequence to the L transformer decoders. First add position embedding for the whole sequence with a rotary embedding way in every decoder. Then perform mask multi-head self-attention for every local window sequence.
5) Get Last Token Embedding
Get the last pix token embedding from every local window pix token embedding sequence.
6) Unified Pix Token And Word Token Mask Multi-head Self-Attention
We integrate word token embedding sequence and pix token embedding sequence into one sequence, and then send to the I transformer decoders to perform mask multi-head self-attention.
7) MLP Head
The final output goes through a MLP head and we get the predict token.

A simple example of the above process is shown in the Figure 5. We will explain in detail the key steps of the above process.

### 3.3.1 Global to Local Window

Assuming that the shape of pix token embedding sequence obtained from image (3, H, W) according to the method in section 3.1.1 is (N,d), N=H*W is the sum of all pixels in the image. We divide the image into (N/K)**2 windows in order from left to right and top to bottom and each window the height and width are both K. The pix token embedding of pixels in each window forms a pix token embeddings sequence. The pix token embedding sequence of all window forms a batch. So the pix token embedding sequence dimension changes from (N, d) to (N/K**2, K**2, d).

In this way, The global pix token embedding sequence becoming a batch local window pix token embedding sequence.

Taking Figure 5 as an example to illustrate: The global pix token embedding sequence changes from (16, d) to (4, 4, d). Both H and W are 4, N is 16 and K is 2.

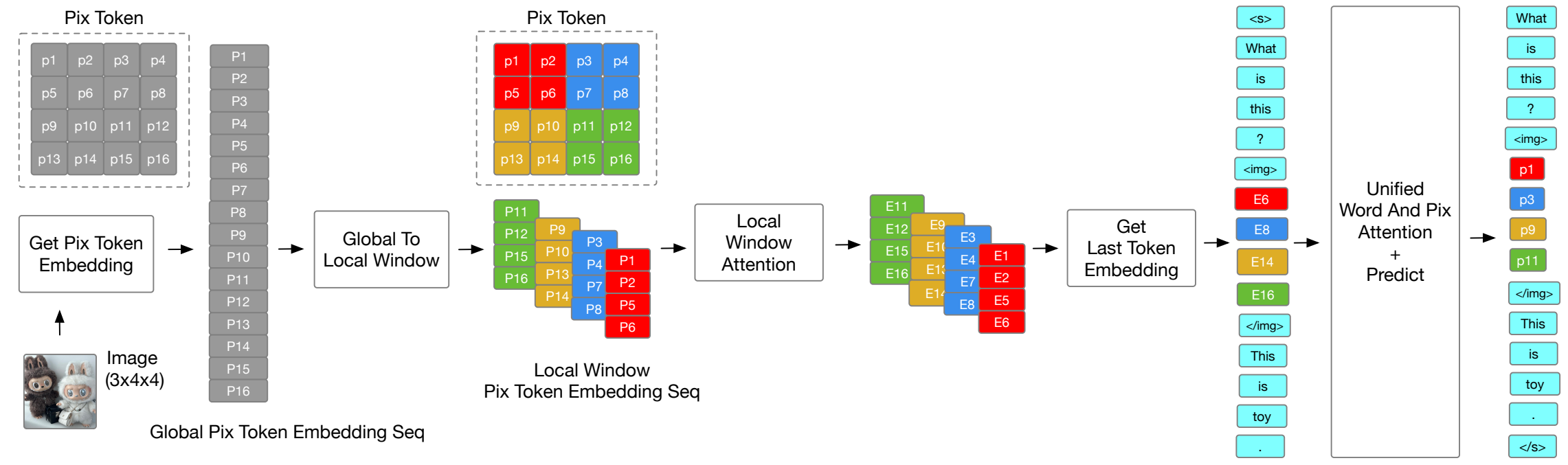

Figure 5: Unified Pix Token And Word Token Model Process.

Note: W must can be divided by the K. If the H cannot be divided by the K, it needs to be filled with pix pad token (section 3.5.1) to the same long as W.

### 3.3.2 Local Window Conditional Attention

We received a batch of local window pix token embedding sequence on section 3.3.1, dimension is $(N/K**2, K**2, d)$. Then perform mask multi-head self-attention operation on this batch sequence.

Position Embedding:
We change the local window pix token embedding sequence to global sequence and add position embedding with a rotary embedding way. After that we change to local window pix token embedding sequence again.

Local Window Conditional Attention:
Perform mask multi-head self-attention operation on K**2 pix token embeddings according to N/K**2 batches.

To illustrate with Figure 5:
The first window, perform mask multi-head self-attention on the sequence $(P_1, P_2, P_5, P_6)$.
The secnd window, perform mask multi-head self-attention on the sequence $(P_3, P_4, P_7, P_8)$.
The third window, perform mask multi-head self-attention on the sequence $(P_9, P_{10}, P_{13}, P_{14})$.
The forth window, perform mask multi-head self-attention on the sequence $(P_{11}, P_{12}, P_{15}, P_{16})$.

### 3.3.3 Get Last Token Embedding

We get the last pix token embedding from every local window of the final output after local window attention. The output dimension is (N/K**2, d).

To illustrate with Figure 5:
We get a token embedding sequence $(E_6, E_8, E_{14}, E_{16})$.
The probabilitie of $E_6$ is $P(E_6|P_1, P_2, P_5, P_6)$.
The probabilitie of $E_8$ is $P(E_8|P_3, P_4, P_7, P_8)$.
The probabilitie of $E_{14}$ is $P(E_{14}|P_9, P_{10}, P_{13}, P_{14})$.
The probabilitie of $E_{16}$ is $P(E_{16}|P_{11}, P_{12}, P_{15}, P_{16})$.

### 3.3.4 Unified Pix Token and Word Token Conditional Attention And Token Predict

**Unified Pix Token and Word Token Mask Multi-head Self-Attention:**
We integrate word token embedding sequence and pix token embedding sequence into one sequence, and then send to the I transformer decoders to perform mask multi-head self-attention.

**Token Predict:**
To predict word token the same as the current method that is next word token. To predict pix token, the model predict the first pix token of the next local window.

**Global Conditional Attention Approximation:**
Observing the entire process, we use a clever method to approximate global conditional attention. The reasons are as follows, refer to Figure 4:
Assuming the input text word token is $X$, the input image pix tokens is $(p1, p2, ..., pN)$, and the output response of model is $Y$. The probability of $Y$ is $P(Y|X, p_1, p_2, ..., p_N)$. But due to the length of pix tokens being too long, for example the length of pix tokens of image with a height and width of 256 is 65556, It's not feasible to do self-attention operation. Our method can avoid this problem, and approximate the goal. We know $(E_k, E_{2k}, ..., E_N)$ are the token embedding after local window conditional attention. The probability of $E_k$ is $P(E_k|p_1, p_2, ..., p_K)$. The probability of $E_{2k}$ is $P(E_{2k}|p_{K+1}, p_{K+2}, ..., p_{2K})$. The probability of $E_N$ is $P(E_N|p_{N-K+1}, p_{N-K+2}, ..., p_N)$. The probability of $Y$ is $P(Y|X, E_k, E_{2k}, ..., E_N)$. we obtain the equation: $P(Y|X, E_k, E_{2k}, ..., E_N) \approx P(Y|X, p_1, p_2, ..., p_K, p_{K+1}, p_{K+2}, ..., p_{2K}, ..., p_{N-K+1}, p_{N-K+2}, ..., p_N) = P(Y|X, p_1, p_2, ..., p_N)$. So our method can approximate the global conditional attention.

We can also take a look at the example in Figure 5.

### 3.4 The advantages of unified token model

#### 3.4.1 Every pix has its true token embedding

ViT forms a fake token embedding by patching the pixels in the image. But this fake token embedding does not have truly adjustable parameters. Pixes do not have their own independence token embedding.

Our model every pix has its own independence token embedding. These pix token embedding has adjustable parameters just like word token embedding. So there is a stronger expressive ability.

#### 3.4.2 The unity of pix token and word token

Our model finally unifies pix token and word token. The number of word Tokens in the current popular generative language model is approximately tens of thousands. There are also tens of thousands of pix tokens in our model, according to different fold factor. We can feel the beauty of this symmetry.

In the unified token space, we can do self-attention operation and token predict for both pix token and word token. But in Llava, we cannot do pix token predict. This is not perfect.

#### 3.4.3 Say bye bye to "the slightest nudge causes the widest chain reaction"

We know that image is encoded as features by ViT in CLIP method. But if a small change is made for the image, for example the license plate number of car in the image changes from 1 to 6, the all floating-point numbers in the encoded features will change. The slightest nudge causes the widest chain reaction, which is not good.

Our model can avoid such problems. Only one or a few token embedding have undergone changes, but most of the others will not change. This makes it easier for model to capture small changes.

#### 3.4.4 Image unsupervised pretraining

GPT-3 tells us that extensive word token unsupervised pretraining is the cornerstone for LLMs to achieve success. But CLIP method is using image supervised training. This is clearly the reason for the missing distribution.

Our model finally makes image unsupervised pretraining possible. Our model can undergo the following pretraining:
**1) Only word token pretraining**
This is the same as the currently popular method. We won't go into detail anymore.

**2) Only pix token pretraining**
We make the first pix token predict of next local window for the input image. Assuming the image starting and ending special tokens is <img> and </img>.
Pix token predict sequence is $(p_1, p_{k+1}, p_{2k+1}, ..., p_{ik+1},$ </img>$)$
The $p_1$ token probability is: $P(p_1|$<img>$)$.
The $p_{k+1}$ token probability is: $P(p_{k+1}|$<img>$, E_k)$.
The $p_{2k+1}$ token probability is: $P(p_{2k+1}|$<img>$, E_k, E_{2k})$
...
The $p_{ik+1}$ token probability is: $P(p_{ik+1}|$<img>$, E_k, E_{2k}, E_{ik})$
The </img> token probability is: $P($</img>$|$<img>$, E_k, E_{2k}, ..., E_{N/K**2})$
$p_{ik+1}$ is the first pix token of the $i + 1$ local window, $i = [0, N/K**2]$.

**3) Mixed word token and pix token pretraining**
There are massive mixed samples of text and image in reality world, for example, image news on the Internet. We can now use these mixed sample for pretraining. This is something that previous methods like Llava were unable to achieve!

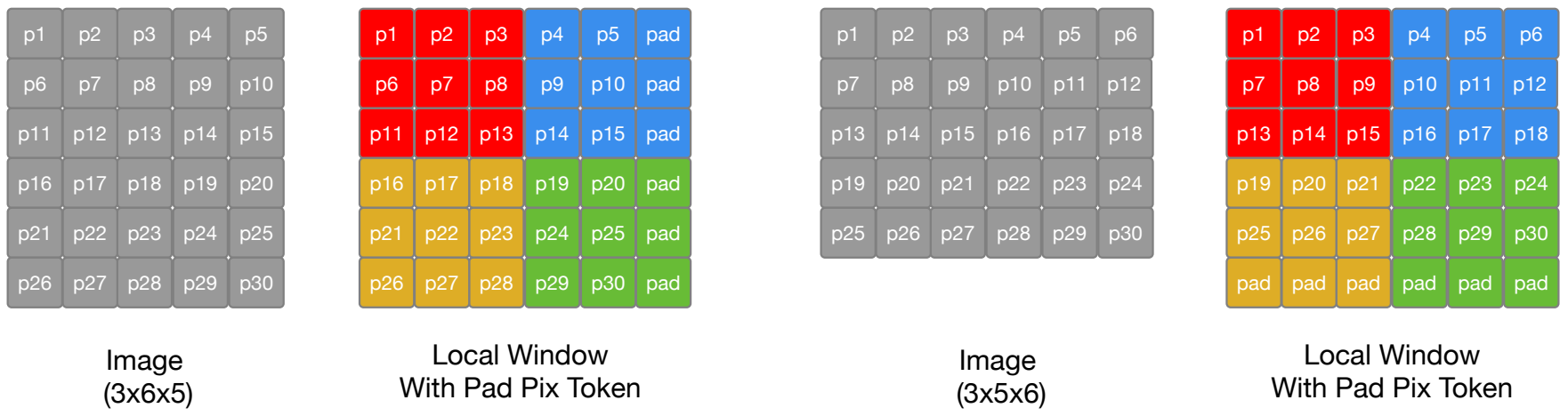

Figure 6: Add Pad Pix Token To Source Image.

But how to deal with batch datas in pretraining? Our model encapsulate the operations (steps 2, 3, 4, and 5 in section 3.3) on image into a module called ImageBlock. It accepts input from batch images and its output is batch pix token embedding sequence.

Taking 3 samples as an example:
sample S1 is only text: (T). sample S2 is text and image mixture: (I, T). sample S3 is text and image mixture: (T1, I, T2). T is a text sequence, I is a shape of (3,H,W) image.

Brief pretraining process:
1) First obtain T of all samples and get word token embedding sequence and label tokens.
2) Second obtain I of all samples and its label tokens, then send it to the ImageBlock to obtain pix token embedding sequence.
3) Connect word token embedding sequence and pix token embedding sequence into a sequence. Connect label tokens of T and I into a sequence.
4) Perform unified conditional attention operation on the connected token embedding sequence and final output predict tokens.
5) Finally, the obtained predict tokens and label tokens will undergo Cross-Entropy Loss operation.

The mixed samples format supported by our model see Appendix A. Users should convert their mixed image and text data into this format.

### 3.5 Implement Supplementation

These are some supplements to the implementation of model details.

#### 3.5.1 Pad Pix Token

During pretraining, batch processing of images is required. Our model supports images of any ratio. To achieve this goal, We set the length and width of the target image both to image size L. we scale the length or width of the source images to L, and add a pad pix token to the remaining shortfall. This pad pix token will not participate in conditional self-attention operations in local window, which is achieved through mask.

As shown in Figure 6. Assuming image size L is 6, local window size is 3. For image (3x6x5), due to insufficient width and the addition of pax pix token, it turns into image (3x6x6). For image (3x5x6), due to insufficient height and the addition of pax pix token, it turns into image (3x6x6).

#### 3.5.2 Sub Local Window

When our model supports large image size L, if local window size is too small, the resulting pix token embedding sequence will be too long. And if local window size is too large, the conditional self-attention operation in local window will take too long time. So we use sub local window attention to solve this problem.

**Specific method:**
Stage one is to use a suitable local window size to divide the global pix token embedding sequence of the source image. After getting the last token embedding, we obtained a pix token embedding sequence.
Stage two continue to perform a smaller local window size on the pix token embedding sequence and carry out conditional self-attention operations in local window.
The subsequent steps are the same as before.

As shown in Figure 7. Assuming image size L is 8, local window size is 2. After stage one local window attention, we got a pix token embedding sequence with a length of 16. And then continue to perform a smaller local window size of 2 on it and got 4 local window pix token embedding sequence. Subsequent local window attention can be carried out on these local sequence and finally obtaining a ix token embedding sequence with a length of 4.

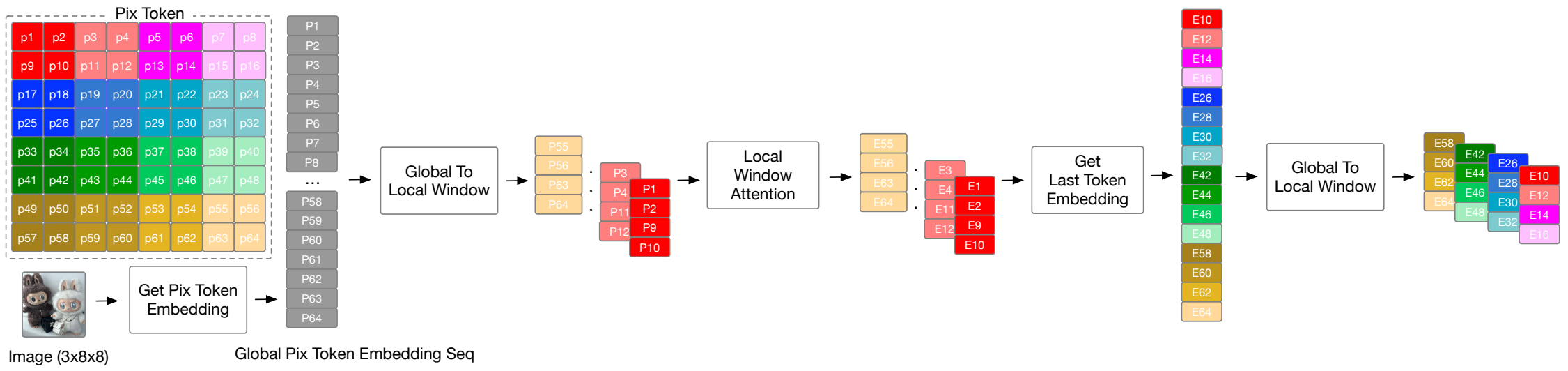

Figure 7: Sub Local Window Attention.

# 4 Experiments

We conducted a only image unsupervised pretraining experiments using our unified pix token and word token model. The purpose of the experiment is to demonstrate the feasibility of our model, and we expect its results to be similar to those of only text unsupervised pretraining like GPT-3.

Due to our lack of computing resources, we are unable to conduct other experiments, such as mixed text and image unsupervised pretraining. We hope everyone can understand. So we conducted the experiment that best demonstrated the potential of our model, that is only image unsupervised pretraining.

## 4.1 Settings

**1) Dataset**
Training dataset: Due to our lack of computing resources, we used a dataset with a small quantity and low resolution. We used 120000 image samples from the llava-cc3m-pretrain-595k dataset, about 6 billion pix tokens.

Validate dataset: We used one thousand samples from the llava-cc3m-pretrain-595k dataset.

**2) Model**
Due to our lack of computing resources, we used a model with 120 millions parameters. It is structurally simple and does not contain the current complex features such as MoE, model parallelism and tensor parallelism and so on. The configuration information of the model is as Table 1.

**3) Training**
We trained our model, using learning rate 0.0006, cosine learning rate scheduler, AdamW with $\beta1 = 0.9$, $\beta2 = 0.999$, dropout of 0.1, weight decay of 0.01, gradient clipping at global norm 1, batch size 2. Image related parameters: image size is 224, color fold factor is 16, and local window size is 16.

## 4.2 Experimental Results

The training loss curve and validation loss curve are shown in the Figure 8. The training loss curve converges with the number of elapsed tokens, which is similar to the only word token pretraining of GPT-3. But compared to pretraining of GPT-3, it converges faster in a relatively small number of pix tokens. The validation loss curve also converges like the training loss curve. This proves the feasibility of our unified token model.

**Experimental conclusion:**
Our model with only 120 millions parameters can achieve good results with limited training data. The effect should be better if we use a tens of billions parameters model. If more massive training data is used, model will inevitably grasp the patterns of the pix token sequence. In short, we believe that when we increase the number of parameters and datas, our model also conforms to the scaling law of generative language model.

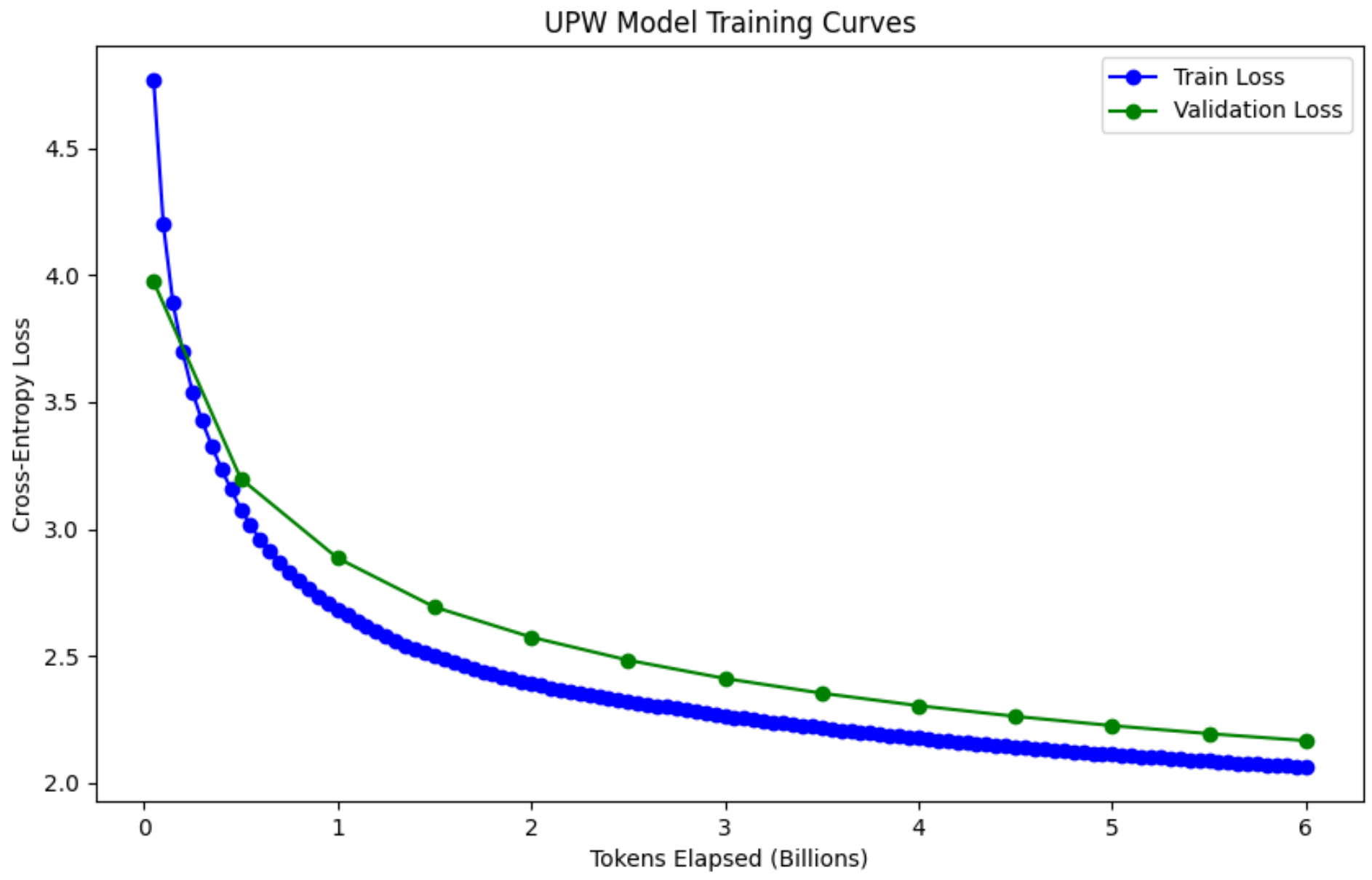

Figure 8: Only Image Unsupervised Pretraining Curvs.

## 5 Conclusion

We first analyzed the limitations of current state-of-art open source generative language models in visual understanding. Then we propose a new model to solve this issues: Unified pix token and word token generative language model. Our model have many advantages, including the ability to undergo image unsupervised pretraining.

We conducted only image unsupervised pretraining experiments using our model. The experimental results have proven that after a certain amount of data pretraining, our model can learn the patterns of the pix token sequence. We believe that when we increase the number of parameters and datas, our model also conforms to the scaling law of generative language model.

The new model we proposed is a perfect way to unify pix token and word token together into generative language model. We believe it can replace the CLIP method currently used in production.

# Appendix

## A Mixed files format

The mixed file contains sentence paragraphs and image references. And request to image referenced using <|image|> and <|/image|> special tokens.
Example 1:

> Provide a brief description of the given image. <|image|>GCC_train_002582585.jpg<|/image|> olive oil is a healthy ingredient used liberally .

Example 2:

> I have two cute pets.
> <|image|>my_dog.jpg<|/image|> This is my dog. It has round, warm eyes that glint like melted honey when it tilts its little head. It wags its curly tail so fast that its whole little butt wiggles when you call its name.
> <|image|>my_cat.jpg<|/image|> This is my cat. It curls up on the windowsill, curling its tail around its tiny paws. It rubs its little head against my hand, purring a soft sleepy rumble.